\documentclass[10pt,twocolumn,letterpaper]{article}

\usepackage{cvpr}              % To produce the CAMERA-READY version
\usepackage{graphicx}
\usepackage{amsmath}
\usepackage{amssymb}
\usepackage{booktabs}

\usepackage[accsupp]{axessibility}  % Improves PDF readability for those with disabilities.

\usepackage{multirow}
\usepackage{multicol}

\usepackage{color, colortbl}
\definecolor{Gray}{gray}{0.9}
\definecolor{LightCyan}{rgb}{0.88,1,1}
\usepackage[first=0,last=9]{lcg}

\definecolor{LightBlue}{rgb}{0.90,0.95,1}
\usepackage[pagebackref,breaklinks,colorlinks]{hyperref}

\def\eg{\emph{e.g}\onedot} 
\def\ie{\emph{i.e}\onedot}

\usepackage[capitalize]{cleveref}
\crefname{section}{Sec.}{Secs.}
\Crefname{section}{Section}{Sections}
\Crefname{table}{Table}{Tables}
\crefname{table}{Tab.}{Tabs.}

\def\cvprPaperID{6314} % *** Enter the CVPR Paper ID here
\def\confName{CVPR}
\def\confYear{2022}

\begin{document}
	
	%%%%%%%%% TITLE - PLEASE UPDATE
	\title{Learnable Irrelevant Modality Dropout for Multimodal Action Recognition on Modality-Specific Annotated Videos}

	\author{Saghir Alfasly$^{1,2}$ \:\:\:\:\: Jian Lu$^{1,3,}$\thanks{Corresponding author}\;  \:\:\:\:\: Chen Xu$^{1,2}$ \:\:\:\:\: Yuru Zou$^{1}$ \\
		%\small Shenzhen University\\
		\small $^{1}$Shenzhen Key Laboratory of Advanced Machine Learning and Applications, Shenzhen University, China\\
		\small $^{2}$Guangdong Key Laboratory of Intelligent Information Processing, Shenzhen, China \:\:\: \small $^{3}$Pazhou Lab, Guangzhou, China\\
		%	 \small Shenzhen Key Laboratory of Advanced Machine Learning and Applications, College of Mathematics and Statistics, Shenzhen University\\
		{\tt\small \{saghiralfasly, jianlu, yuruzou\}@szu.edu.cn}, {\tt\small chenxuszu@sina.com}
	}

	\maketitle
	
	%%%%%%%%% ABSTRACT
	\begin{abstract}
		With the assumption that a video dataset is multimodality annotated in which auditory and visual modalities both are  labeled or class-relevant, current multimodal methods apply modality fusion or cross-modality attention. However, effectively leveraging the audio modality in vision-specific annotated videos for action recognition is of particular challenge.
		
		To tackle this challenge, we propose a novel audio-visual framework that effectively leverages the audio modality in any solely vision-specific annotated dataset. We adopt the language models (e.g., BERT) to build a semantic audio-video label dictionary (SAVLD) that maps each video label to its most K-relevant audio labels in which SAVLD serves as a bridge between audio and video datasets. Then, SAVLD along with a pretrained audio multi-label model are used to estimate the audio-visual modality relevance during the training phase. Accordingly, a novel learnable irrelevant modality dropout (IMD) is proposed to completely drop out the irrelevant audio modality and fuse only the relevant modalities. Moreover, we present a new two-stream video Transformer for efficiently modeling the visual modalities. Results on several vision-specific annotated datasets including Kinetics400 and UCF-101 validated our framework as it outperforms most relevant action recognition methods.
		
	\end{abstract}
	
	\begin{figure}
		\centering
		\includegraphics[trim=0 0 0 0, width=8.5cm,clip, keepaspectratio]{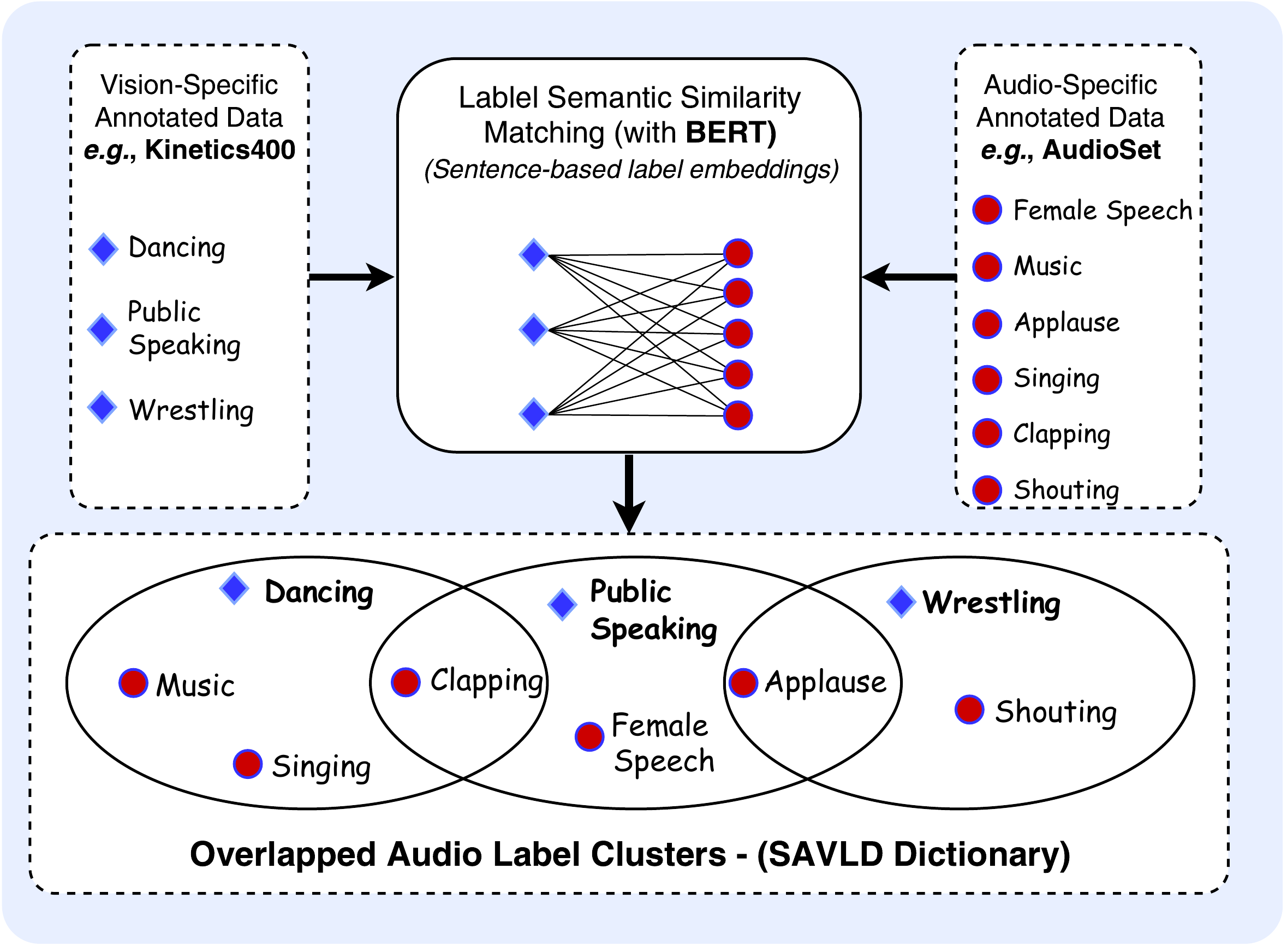}
		\caption{\textbf{Conceptual Overview:} Vision-Audio Label Mapping. In an overlapped manner, our method performs a cross-dataset textual labels mapping in which audio labels are mapped to their most closely relevant video-based human activity labels by using language models, \eg, BERT. The resulting clusters compose a SAVLD dictionary that serves as a bridge between video and audio datasets. Since our framework trains human activity multimodal models on vision-specific datasets, we use SAVLD to leverage the auditory modality in videos.}
		\label{fig:ConceptualOverview}
	\end{figure}
	
	%%%%%%%%% BODY TEXT
	\section{Introduction}
	\label{sec:intro}
	One of the deep neural network (DNN) learning schemes to improve video understanding is to leverage as many input modalities as available, such as audio, RGB frames, motion, textual data, the visible text on videos, and human skeleton joints. Therefore, multimodal learning has shown a remarkable improvement in video-based action recognition. These methods process either each modality with an independent DNN or all with a single shared DNN. The modalities or their feature representations are fused either in an early fusion, middle fusion, or late fusion manner \cite{BottleneckFusion,fusion,EfficientMultiModalTransformer}. Another fusion scheme is proposed by adopting cross-modality attention or gated units \cite{twostreamMulitmodal,MultimodalCompressedVideo,GatedUnit,UnimodalBetterMultimodal}.
	
	\noindent\textbf{Motivation.} Most of the attention and fusion methods \cite{fusion} boost the audio-visual models' performance on audio-vision correspondence-based videos in which auditory and visual modalities are corresponding or at least relevant as in Kinetics-Sounds \cite{AVC} and EPIC-Kitchen dataset \cite{EPIC-KITCHENS}. However, these fusion methods do not provide similar performance boost on the vision-specific annotated datasets, such as Kinetics400 \cite{kinetics400} and UCF-101 \cite{ucf101}. This condition is due to the high irrelevance and non-correspondence between visual and auditory modalities as concluded in \cite{BottleneckFusion}. In this case, multimodal methods do not effectively gain maximum benefit from unlabeld audio modality with noisy relevance on vision-specific datasets. For example, {\tt\small music} can largely indicate that the human activity is {\tt\small dancing}. However, music audio can be associated with several other human activities, such as car driving, or simply can be found as background audio in any edited video. This interprets the findings in an interesting study \cite{UnimodalBetterMultimodal}, which showed that unimodal models consistently outperform multimodal DNN in modality-specific datasets because multimodal networks are more prone to overfitting with their increase in capacity. This finding motivated us to look for a method that leverages the relevant audio modality while completely dropping out the irrelevant modality in training and inference phases.    
	
\begin{figure*}
		\centering
		\includegraphics[trim=0 0 0 0, width=17.5cm,clip, keepaspectratio]{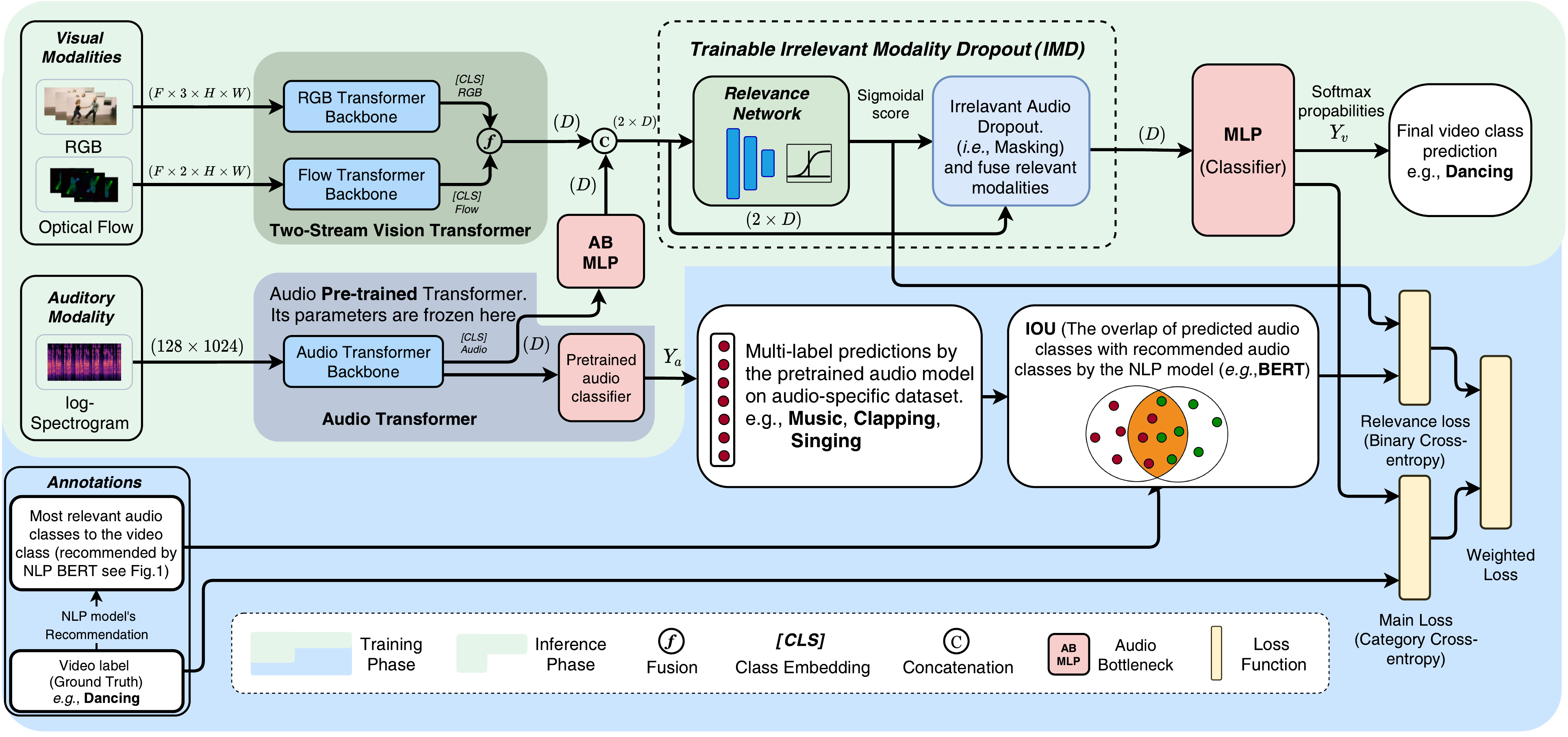}
		%	\vspace*{-3mm}
		\caption{\textbf{Proposed Multimodal Learning Framework}. It uses three modalities: RGB, flow, and audio. Audio modality features are obtained by an off-the-shelf Transformer trained on the audio dataset AudioSet. Here, audio Transformer parameters are not trainable except the added AB which works as a bridge between our trainable network and the pre-trained frozen audio model. The trainable IMD is the main part of this framework, it estimates the auditory-visual modality relevance of the input video using the RN, and then it decides whether to fuse the audio modality or to completely drop it out using thresholding and masking layers (Fig. \ref{fig:IMD}). IMD is optimized with binary cross-entropy. The IOU is computed between the audio Transformer multi-label predictions and the corresponding $k$ relevant audio labels generated by SAVLD. The IOU is normalized into the [0–1] range to match the relevant network (RN) output range.}
		\label{fig:framework}
	\end{figure*}

	\noindent\textbf{Contributions.} To tackle the aforementioned challenge, we present a novel multimodal training framework that trains action recognition networks with the best audio-visual modality combination on visual modality-specific datasets. In our method, we automatically estimate the audio-visual modalities relevance by leveraging the pretrained audio classification models on large audio-specific datasets such as AudioSet \cite{AudioSet} and VGGSound \cite{VGGsound}. This task is effectively achieved in two phases. In the first phase, we map audio labels of an audio-specific dataset to their most semantically relevant labels in the vision-specific dataset. This process is achieved using the language pretrained models, \eg, BERT \cite{BERT} or GloVe \cite{GloVe}. After obtaining the semantic sentence-based embedding of each label, we perform cross-dataset label matching and generate overlapped label clusters each of which contains a single label from the video dataset along with a set of its most semantically similar labels in the audio dataset as shown in Fig. \ref{fig:ConceptualOverview}. For example, {\tt\small Dancing} label in Kinetics400 and {\tt\small Music}, {\tt\small Singing}, and {\tt\small Clapping} labels in AudioSet are in the same cluster. Overall, the resulting overlapped clusters compose a semantic audio-video label dictionary (SAVLD) that serves as a bridge between video and audio datasets for the training supervision in our next phase.
	In the second phase, as a transfer learning task, our framework adopts a pretrained audio Transformer to generate highly semantic audio features and multi-label audio predictions. The audio dataset used for building the SAVLD should be the same dataset on which the audio Transformer has been trained. Overall, the outputs of the audio Transformer and the SAVLD are then used by our proposed framework to annotate audio, as depicted in Fig.\ref{fig:framework}.
	
	After obtaining the audio modality predictions using any audio pretrained model (\eg, AST), our framework guided by the SAVLD drops the irrelevant audio modality. This is performed by using a novel trainable irrelevant modality dropout (IMD) that consists of two main modules: The first module is a neural network termed relevance network (RN) that receives auditory and visual modalities and decides whether they are relevant. The output of this network is sigmoidal predictions representing the relevance level that is used by the second module to decide whether to fuse the two modalities for the final video classification or to drop the audio modality. We further improve our framework learning by proposing a new intra-class cross-modality augmentation in which it randomly pairs auditory and visual modalities of the same class from different videos. This augmentation method may show a negative effect on some applications of audio-visual networks when the audio and video are required to be corresponding as the case of speech recognition. However, in our case, the audio modality is required to be relevant but is not necessarily to be aligned or accurately correspond to the visual modality because we tackle the problem of human activity recognition in this work. This motivates us to propose this augmentation method that empirically provides a reasonable performance boost. 
	
	Generally, our framework is an entirely convolution-free Transformer-based network. It leverages three video modalities: RGB frames, optical flow, and audio. The RGB and optical flow modalities are processed by using our efficient two-stream video Transformer, whereas the audio modality is processed by any audio pretrained Transformer. 
	
	\textbf{The key contributions of this work are as follows}:
	\vspace*{-1mm}
	\begin{itemize}
		\item A novel multimodal human activity recognition framework that leverages the power of the NLP BERT model and the pretrained audio classification models is proposed to automatically annotate audio modality. It effectively trains audio-visual action recognition models on any vision-specific annotated dataset.
		\vspace*{-2mm}
		\item A novel learnable IMD network is proposed to completely drop out the irrelevant audio modality, whereas the relevant modalities are fused on the basis of their relevance level. 
		\vspace*{-2mm}
		\item An efficient two-stream video Transformer is designed to learn the visual modality with few parameters as compared to the relevant video Transformers.
		\vspace*{-2mm}
		\item An intra-class cross-modality augmentation method is proposed for generating more training samples by allowing each audio modality sample to be paired with any visual modality sample among the same class.
	\end{itemize}
	
	\section{Related Work}
	\label{sec:intro}
	\noindent\textbf{Multimodal Action Recognition.}
	Multimodal action recognition aims to exploit multimodality input for better human activity recognition. DNNs have empowered this learning scheme, which enriches the global feature learning \cite{EfficientMultiModalTransformer,UnimodalBetterMultimodal,GatedUnit,DropToken,BottleneckFusion,MultimodalCompressedVideo}. On the basis of the modality combination, some methods use RGB frames along with motion modality as in the two-stream networks \cite{2streams2014,TwoStream2016Fusion,twostreamMulitmodal}, with the skeleton modality \cite{MMTM}, with detected objects \cite{fusion}, with textual modality \cite{VLM}, or with the audio modality \cite{BottleneckFusion,MMTM,AVC,ObjectsthatSound,hardModalityDropout,fusion2} which is the most common multimodality  approach to learn global representation for video understanding. More than two modalities are exploited by some methods \cite{fusion,UnimodalBetterMultimodal,DropToken,EPICFusion,MultimodalCompressedVideo} to improve action recognition. Another approach is followed for obtaining video modalities as in \cite{MultimodalCompressedVideo,compressedvideo}, which uses the compressed videos as a source of four modalities, I-frames, motion vectors, residuals, and the audio modalities. Using these modalities, the 2D/3D convolution networks have shown substantial progress in global feature modeling  \cite{MMTM,AVC,ObjectsthatSound,fusion,CluesLabels,hardModalityDropout,fusion2,twostreamMulitmodal}. However, multimodal video Transformers \cite{DropToken,BottleneckFusion,EfficientMultiModalTransformer,MultimodalCompressedVideo} have shown a competitive performance as a normal reflection of the remarkable Vision Transformer (ViT) success witnessed on image recognition \cite{ViT}. Although, supervised multimodal learning provides good performance \cite{MMTM,AVC,ObjectsthatSound,fusion,EfficientMultiModalTransformer,UnimodalBetterMultimodal,GatedUnit,BottleneckFusion,MultimodalCompressedVideo}, some methods tend to leverage the huge unlabeled video data to learn multimodal networks in self-/weakly supervised scheme \cite{DropToken,Alayrac2020,CluesLabels,Owens2018}
	
	\noindent\textbf{Modality Fusion.}
	One of the main components of any multimodal algorithm is its modality fusion module that is used to fuse and derive the cross-modality representations for the final prediction. Several fusion methods have been proposed and can be categorized into early, mid, and late fusion. These fusion approaches are adopted and empirically studied in more detail in \cite{TwoStream2016Fusion,BottleneckFusion,fusion,EfficientMultiModalTransformer,UnimodalBetterMultimodal,GatedUnit}. The fusion methods vary between a simply modality feature aggregation, concatenation, and cross-modality attention \cite{BottleneckFusion,CluesLabels}. Recently, several multimodal methods \cite{BottleneckFusion,twostreamMulitmodal,MultimodalCompressedVideo} leverage the attention mechanism to perform cross-modality feature modeling resulting in a remarkable improvement in action recognition. 
	
	\noindent\textbf{Modality Dropout and Gating.}
	Several methods, including the well-known dropout \cite{dropout}, have been proposed to improve the training process of the deep networks, which prevents DNNs from overfitting. Model overfitting becomes more challenging when it comes to multimodal training particularly for video understanding, as concluded in \cite{UnimodalBetterMultimodal}. This is due to the small video dataset size as compared to image-based datasets and due to the bias towards majority classes resulting by highly imbalanced datasets. Therefore, several methods use random modality dropout, which is first used in \cite{ModDrop}, for improving the multimodal gesture recognition training. By contrast, a  drop input token  is proposed in \cite{DropToken} for reducing the training time at the cost of the prediction quality. Other studies are presented in \cite{GatedUnit,gatemodality}, which propose a gated multimodal unit for text-vision fusion in case two modalities are relevant.
	
	\textbf{Our work differs from previous methods in the following aspects.}
	First, our vision model is a convolution-free two-stream Transformer that derives spatiotemporal features with few parameters as compared to \cite{BottleneckFusion,MultimodalCompressedVideo,DropToken}. Second, unlike \cite{DropToken,GatedUnit,gatemodality}, our proposed IMD is learnable that learns to completely drop the irrelevant modality whereas the relevant modalities are passed for the fusion step in both training and inference phase. IMD aims to learn multimodality representations from modality-specific annotated datasets. Thus, the proposed IMD is different in its structure, purpose, and training approach. Third, to the best of our knowledge, our work is the first attempt to exploit the power of both NLP and audio Transformers to automatically annotate audio modality that significantly boosted the action recognition performance. Fourth, the audio-visual correspondence (AVC) \cite{AVC}, which considers the dataset as audio-visual corresponding, learns to align auditory and visual modalities and decides whether they are extracted from the same time track of the same video, on the basis of only positive and negative labels. However, our relevance network (RN) does not learn the modality correspondence or alignment. Instead, it learns to simply decide whether the audio modality is class-relevant to the visual modality or not based on the generated label dictionary SAVLD regardless of which videos are used to sample the input modalities. Finally, our proposed intra-class cross-modality augmentation is different from \cite{mixup}, which is an inter-class sample mixing. It is also different from \cite{CluesLabels}, which exchanges audio and video tracks for one time as a preprocessing step to ensure each input sample involves auditory and visual modalities. 
	
	\begin{table}
		\fontsize{8}{9}\selectfont
		\centering
		\caption{Samples of video labels in Kinetics400 dataset and their most relevant labels in AudioSet dataset selected by KNN after a semantic label embedding mapping with BERT, when $k=5$.}
		\vspace*{-2mm}
		\begin{tabular}{ll} 
			\toprule
			Label & Relevant AudioSet Labels\\
			\toprule
			applauding & speech;applause;whistling;chime;clapping\\
			clean and jerk & fill with liquid;pump liquid;filing rasp;rumble;rustle\\
			feeding birds & wild animals;insect;mosquito;bird;patter\\
			sniffing & whimper;growling;cheering;whispering;rattle \\
			sneezing & gurgling;snoring;babbling;gargling;rapping \\
			tickling & whispering;rustle;cheering;growling;screaming\\
			yawning & babbling;rapping;frying food;gurgling;snoring \\
			writing & writing;speech;typing;chatter;mechanisms \\
			\bottomrule
		\end{tabular}
		\label{tab:SAVDE}
	\end{table}
	\section{Proposed Multimodal Framework}
	\label{sec:method} 
	\subsection{Label Dictionary SAVLD }
	SAVLD is a cross-dataset label dictionary in which each label in an audio-specific dataset is semantically mapped into one or more labels on the vision-specific dataset. We achieved this preprocessing step by using the well-known NLP model BERT \cite{BERT}. We obtain the sentence-based semantic embeddings of each textual label in audio and video datasets because most human activity video and audio labels involve more than one word. Each video label embedding is matched with all audio label embeddings. The most $k$ relevant audio labels are picked by using the k nearest neighbor algorithm. However, the
	matching process does not require much time because the number of labels in two datasets is small, that is, Kinetics400 has 400 class labels and AudioSet has 527 class labels. Thus, finding the closest $k$ similar labels can be done simply by using any distance metric. We obtain a dictionary in which each video label has $k$ audio labels of which they all are considered semantically similar. This dictionary can be considered $U_v$ overlapped clusters, where $U_v$ denotes the number of video dataset classes. The cluster overlap is represented here in terms of audio labels only. A conceptual overview of building SAVLD is illustrated in Fig.\ref{fig:ConceptualOverview}, whereas a set of real Kinetics400-AudioSet mapped labels is shown in Table \ref{tab:SAVDE}.
	\subsection{Visual-Modality Two-Stream Transformer}
	To leverage the pre-trained knowledge on large image datasets \cite{MLSL-imagenet}, we adopt several parts from the image-based ViT \cite{ViT} and the video Transformer \cite{timesformerT} to build our two-stream video Transformer.
	
	\noindent\textbf{Tokenization.} Similar tokenization part of \cite{timesformerT} is used in which the input visual-modality sample $\textbf{V} \in \mathbb{R}^{T \times C \times W \times H}$, of $T$ frames and $C=3$ for RGB modality and $C=2$ for optical flow modality, is tokenized into $T \times N$ patches in which each frame is projected to $N=HW/P^2$ patches each of which has a spatial dimension of $P^2$. After organizing the obtained $T \times N$ patches, each patch is mapped to $\mathbf{x}_{j,i} \in \mathbb{R}^D$, where $j=1,\cdots,T$ , $i=1,\cdots,N$, and $d$ denotes the embedding size to be used throughout the visual two-stream transformer. 
	\vspace*{-2mm}
	\begin{equation}
	\mathbf{z}_{j,i}^0 = \mathbf{x}_{j,i} \mathbf{E} + \mathbf{E}^{pos}_{j,i},
	\label{eq:embedding}
	%\end{aligned}
	\end{equation}
	where for each patch $\mathbf{x}_{j,i}$, a spatiotemporal positional embedding $\mathbf{E}^{pos}_{j,i} \in \mathbb{R}^D$  is added. Additionally, a learnable classification embedding  $\mathbf{z}_{0,0}^0\in \mathbb{R}^{D} $ is added to learn the visual-modality semantic video representation. 
	
	\begin{figure}
		\centering
		\includegraphics[trim=0 0 0 0, width=8.3cm,clip, keepaspectratio]{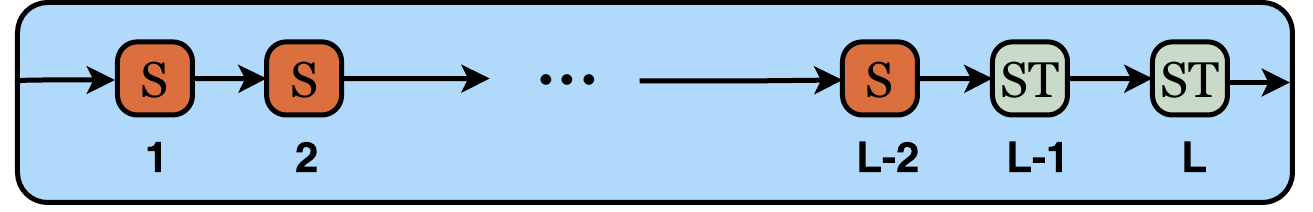}
		%	\vspace*{-3mm}
		\caption{\textbf{Visual-Modality Transformer Encoder.} It consists of $L-2$ Spatial \textbf{S} blocks  and $2$ Spatio-Temporal \textbf{ST} blocks. \textbf{ST}s are spatial–temporal
			factorized-based attention blocks to preserve the Transformer efficiency}
		\label{fig:smartpicker}
	\end{figure}
	\noindent\textbf{Visual-Modality Transformer Encoder}.
	Our visual two-stream Transformer encoder involves $L$ stacked blocks, each of which is a spatial-based module except the last two blocks $\{L,L-1\}$ which involve a spatiotemporal factorized self-attention to better model the spatiotemporal knowledge. Using this approach, our Transformer provides competitive performance with few parameters as compared to \cite{timesformerT} which involves spatiotemporal attention in each block. Each Transformer block consists of a multiheaded self-attention (MSA) \cite{selfAttention}, LayerNorm (LN) \cite{LN}, and multilayer perception (MLP) modules. The first $L-2$ blocks of the proposed Transformer encoder can be formulated as:
	\vspace*{-2mm}
	\begin{equation}
	\begin{aligned}
	\mathbf{z}_{j,i}^\ell = \mathrm{ \large MSA} \left(\mathrm{LN}(\mathbf{z}_{j,i}^{\ell-1}) \right)  + \mathbf{z}_{j,i}^{\ell-1},
	\label{eq:space}
	\end{aligned}
	\end{equation}
	where $\ell \in \{0,...,L-2\}$. The last two blocks can be formulated as:
	\vspace*{-3mm}
	\begin{equation}
	\begin{aligned}
	\grave{\mathbf{z}}_{j,i}^{\ell} =  \mathrm{ \large LR} \left(  \mathrm{ \large MSA}^{time} (\mathrm{ \large LN}(\mathbf{z}_{j,i}^{\ell-1}) ) \right)  + \mathbf{z}_{j,i}^{\ell-1},\\
	\mathbf{z}_{j,i}^{\ell}= \mathrm{ \large MSA}^{space} \left( \mathrm{ \large LN}(\grave{\mathbf{z}}_{j,i}^{\ell} \right) + \grave{\mathbf{z}}_{j,i}^{\ell},
	\label{eq:lastblock}
	\end{aligned}\vspace*{-2mm}
	\end{equation}
	where $\mathrm{ \large LR}$ refers to a linear layer that added after each temporal module. The classification embeddings of RGB and optical flow streams are fused with two fully-connected layers each of which is followed by ReLU activation.
	
	\subsection{Audio Transformer}
	We choose the pretrained AST \cite{AST} because it is the most closely similar Transformer to our video Transformer. We adopt AST to extract high semantically audio features and to perform a multi-label prediction. AST has been built on the basis of ViT \cite{ViT} in which it treats audio modality processing as a vision task because it converts input audio modality into $ 128\times 100t $ spectrogram. AST is mostly identical to ViT except for a few changed Transformer parameters, where AST uses patch size $P=16$ with a stride of $10$. Overall, we follow the audio input normalization and inference settings in \cite{AST} because we adopt AST as a downstream task with frozen parameters. We obtain the output predictions of AST on its audio dataset (\ie, AudioSet) to obtain the video label using the label dictionary SAVLD. Additionally, we obtain the learned class embedding which is then passed to a learnable audio bottleneck.       
	
	\noindent\textbf{Audio Bottleneck (AB)}.
	%We add a learnable audio bottleneck  as a semantic representation bridge between audio and video datasets, such as AudioSet-Kinetics400. This condition is because all AST’s parameters are frozen while training our video-based action recognition model. It consists of one $\mathrm{ \large LN}$ and two $\mathrm{ \large LR}$ each of which is followed by ReLU activation. 
	We add a learnable audio bottleneck as a semantic representation bridge between audio and video datasets, such as AudioSet-Kinetics400. This condition is because all audio model parameters are frozen while training our video-based action recognition model. This bottleneck plays a significant role by transforming the audio features from audio-dataset-oriented predictions to video-oriented predictions in the training phase. Simply, it consists of one $\mathrm{ \large LN}$ and two $\mathrm{ \large LR}$ each of which is followed by ReLU activation. 
	%\subsection{Trainable IMD}
	\subsection{Trainable Irrelevant Modality Dropout (IMD)}
\begin{figure}
		\centering
		\includegraphics[trim=0 0 0 0, width=8.3cm,clip, keepaspectratio]{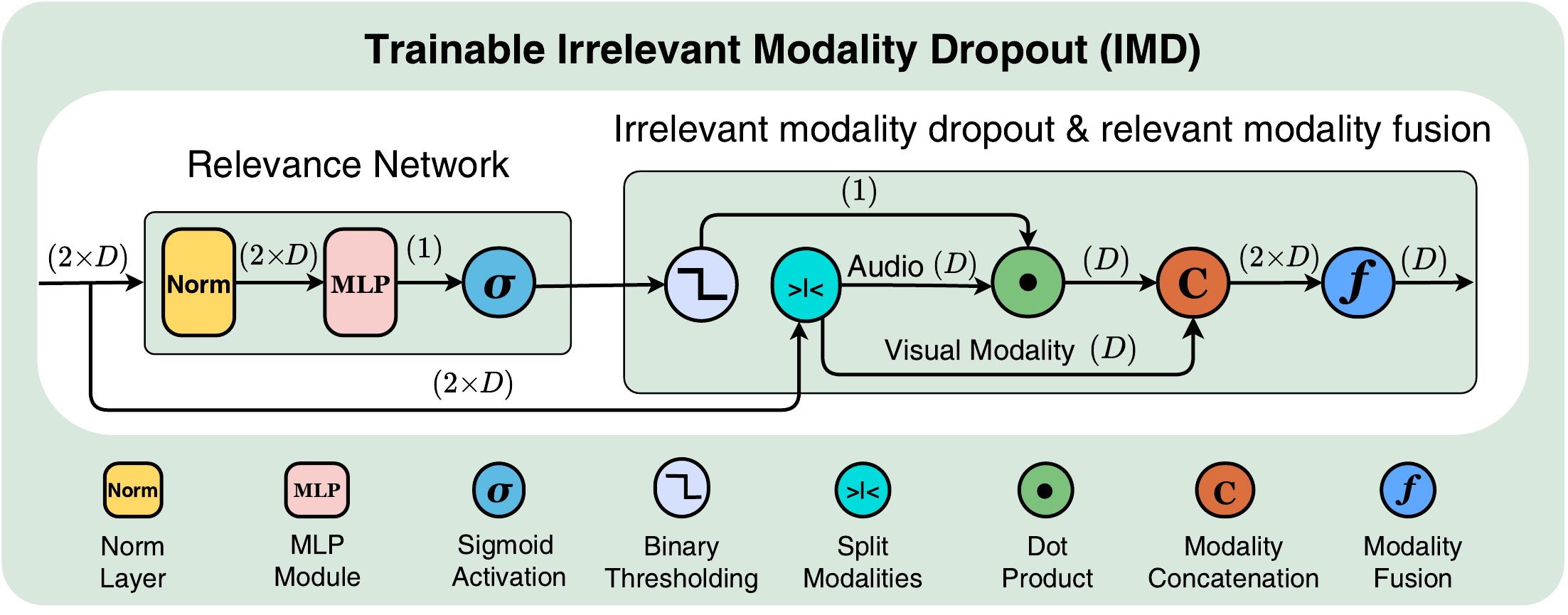}
		%	\vspace*{-3mm}
		\caption{\textbf{Proposed Irrelevant Modality Dropout (IMD)}. It is composed of two main networks. Relevance network (RN) which estimates the class-based audio-vision relevance of the input clip. The IMD network performs two steps: the output of the RN is thresholded and masked as in Eqs. \ref{eq:threshold}–\ref{eq:mask}. Consequently, the audio modality is completely dropped or fused as in Eq. \ref{eq:drop}.}
		\label{fig:IMD}
	\end{figure}

	IMD is the main part of our framework that automatically drops out the irrelevant audio modality. IMD consists of two main modules: The first module is a simple relevance network that receives the concatenated modalities of auditory and visual Transformers and decides whether they are relevant. The output of this network is sigmoidal predictions representing the relevance level. The sigmoidal predictions are forwarded to a binary thresholding layer to decide whether audio-modality embedding can be fused with visual-modality embedding for the final classification or to drop it. The thresholding and dot product layers compose a masking layer in which the irrelevant modality is multiplied with $0$. The proposed IMD can be formulated as follows:
	\vspace*{-1mm}
	\begin{equation}
	\begin{aligned}
	rev = \sigma (\mathrm{ \large MLP} (\mathrm{LN}(\mathbf{z}_{0,0}^{L}))  +\mathbf{z}_{0,0}^{L}),
	\label{eq:RN}
	\end{aligned}
	\end{equation}
	\vspace*{-5mm}
	\begin{equation}
	\begin{aligned}
	\delta = \begin{cases}0,& \text{if } z < \alpha\\rev, & \text{otherwise}\end{cases},
	\label{eq:threshold}
	\end{aligned}
	\end{equation}
	\vspace*{-2mm}
	\begin{equation}
	\begin{aligned}
	{z}_{0,0}^{(a)L},{z}_{0,0}^{(v)L} = split({z}_{0,0}^{L}),
	\label{eq:split}
	\end{aligned}
	\end{equation}
	\vspace*{-3mm}
	\begin{equation}
	\begin{aligned}
	{z}_{0,0}^{L} = Concate(({z}_{0,0}^{(a)L} \cdot \delta),{z}_{0,0}^{(v)L}),
	\label{eq:mask}
	\end{aligned}
	\end{equation}
	\vspace*{-3mm}
	\begin{equation}
	\begin{aligned}
	z_{av}= \mathrm{ \large MLP} (\mathrm{LN}(\mathbf {z}_{0,0}^{L})),
	\label{eq:drop}
	\end{aligned}
	\end{equation}
	where $\alpha$ denotes the defined threshold for the audio-visual modality relevance and $z_{av}$ is the fused output. Figure \ref{fig:IMD} illustrates the IMD network and its main layers. Although it is designed for audio-modality dropout, it may be adopted for other modalities in the case of textual-vision applications whenever possible to easily build a cross-dataset label dictionary as the SAVLD we built for audio-visual training.
	\vspace*{-4mm}
	\subsection{Framework Optimization}
	Our framework has two learning targets; optimizing the modality Transformers to learn the final human activity and optimizing the relevance network to accurately estimate the audio-video relevance. Therefore, we adopt two cross-entropy losses. The first loss is a binary cross-entropy for optimizing the RN by using the label dictionary SAVLD. The second loss is the default category cross-entropy for the classification learning, which is weighted with the RN loss to form the final loss. Before the category cross-entropy is applied to the output fused class embedding, it is first fed into the final classification MLP as follows:
	%\vspace*{-2mm}
	\begin{equation}
	\begin{aligned}
	Y_v = Softmax(\mathrm{ \large MLP} (\mathrm{LN}(z_{av})))
	\label{eq:mlp}
	\end{aligned}
	\end{equation}
	\vspace*{-5mm}
	\subsection{Intra-Class Cross-Modality Augmentation}
	We propose an intra-class cross-modality augmentation for human activity recognition, which has shown a positive impact in the Transformer training because human activity recognition on video does not require audio-video correspondence as in the case of speech-visual recognition. Thus, in our work, we focus on relevance rather than correspondence in which the audio modality should be relevant to the visual class label regardless of its video source. Therefore, we run a cross-modality intra-class augmentation by pairing audio and visual modalities from different videos but belong to the same class.

	\section{Experimental Results}
	\subsection{Experiments Setup}
	\noindent\textbf{Datasets and Evaluation Metrics.} Since our aim is to tackle the problem of multimodal video understanding in vision-specific datasets, we have conducted an extensive experiments on two well-known video datasets. \textbf{Kinetics400} \cite{kinetics400} is downloadable from YouTube, where we used in this work $241,722$ videos of $400$ classes for training, and $19,877$ videos for validation. The videos of Kinetics400 are trimmed in $ 10 $ seconds. 
	\textbf{UCF-101} \cite{ucf101} contains $13,320$ videos of $101$ classes. Its video length about $7.2$ seconds. It is split into three training/validation splits.
	We adopted AST audio model pretrained on \textbf{AudioSet} \cite{AudioSet}, which is considered one of the largest available audio dataset. It contains $ 1.8 $M sound clips of $ 10 $-second length. We followed the most used evaluation metric Top-$1$ and Top-$5$. We sampled $3$ spatial crops and $4$ temporal clips from each video and averaged their obtained accuracy. We compared our multimodal video Transformer with state-of-the-art methods in terms of computational cost floating-point operation and the number of parameters
	
	\noindent\textbf{Implementation Details.}
	The entire proposed framework is built in Pytorch \cite{pytorchh}. The training and evaluation phases are conducted on two machines (\ie, $8$ $\times$ TITAN RTX and $2$ $\times$ RTX $2080$ Ti). Unless otherwise specified, we ensure to run each set of the framework training instances by using the same settings on the same machine. The GPU's memory is saved for the visual RGB and flow stream because AST is frozen during training and inference, except its bottleneck AB. The input RGB modality is sampled as a clip with spatial resolution of $3\times 224\times 224$ and temporal length of $8$-frames. Similarly, flow modality is sampled with the same RGB spatial and temporal settings except for the channel of $2$ to represent the $x$ and $y$ flow components. We trained each instance with a global input batch of $64$ for $15$ epochs. SGD optimizer is utilized with an initial learning rate of $0.005$ and weight decay of $1e-4$. Our visual video Transformer uses a patch of $16\times 16$ size, block depth $12$, embedding size $D=768$, and self-attention of $12$ heads. The label dictionary SAVLD is saved as a text file, where each row represents the video dataset label and its $k$ relevant audio labels. 
	\subsection{Ablation Study}
	In this section, we conduct a set of extensive ablation studies on the proposed framework variants to explore their optimal settings. We start with SAVLD building techniques and then, the modality combination on the visual-modality specific dataset performance. Finally, the effect of IMD elements is studied. \vspace*{-3mm}
	\subsubsection{Semantic Audio-Video Dictionary Quality}
	\noindent\textbf{Embedding Extractor.} We chose the well-known NLP model termed BERT to extract the best sentence-based semantic features for each textual label due to two main reasons: first, BERT has been trained in an unsupervised manner and has undergone millions of unorganized text from real-world sources. Thus, it can provide better semantic features that can detect the most relevant text. Secondly, unlike word2vec \cite{word2vec} and GloVe \cite{GloVe}, BERT as a text feature extractor is a context-aware text analyzer that provides different embedding for the same word depending on its position in the sentence. The performance of the proposed framework on Kinetics400 using SAVLD constructed with the three embedding extractors is reported in Table \ref{tab:SAVD}.
	\begin{table}
		\fontsize{9}{11}\selectfont
		\centering
		\caption{Impact of embedding extractor and similarity distance methods on the AudioSet-Kinetics400 SAVLD quality.}
		\vspace*{-3mm}
		\begin{tabular}{cccc} 
			\bottomrule
			%		\toprule
			Embed. Extractor & Similarity & Top-1 (\%) & Top-5 (\%)\\
			\toprule
			\textbf{word2vec} &Euclidean&$ 80.9 $&$ 94.3 $\\
			\textbf{GloVe} &Euclidean&$ 81.2 $&$ 94.5 $\\
			\textbf{BERT} &Euclidean&$ 82.8 $&$ 95.7 $\\
			\hline
			BERT &\textbf{Manhattan}&$ 82.6 $&$ 95.5 $\\
			BERT &\textbf{Cosine}&$ 82.8 $&$ 95.9 $\\
			\toprule
		\end{tabular}
		\label{tab:SAVD}
	\end{table}
	
	\noindent\textbf{Embedding Matching.} Three vector similarity matching methods, namely, Euclidean, Manhattan, and cosine similarity distance are evaluated. The lower part of Table \ref{tab:SAVD} reports their performance on BERT embedding matching. Notably, the similarity distance function does not show large effect on the framework performance.\vspace*{-3mm}
	
	\subsubsection{Modality Combination}
	Modality combination on human activity recognition with the proposed framework is evaluated. Table \ref{tab:Modality} reports three modality combination schemes, namely, unimodal, bimodal, and multimodal on all the possible three modalities, audio, RGB, and optical flow. The use of more than one modality can remarkably improve video-based action recognition However, RGB remains the modality with the largest effect on recognition accuracy.
	\begin{table}[!h]
		\fontsize{9}{12}\selectfont
		\centering
		
		\caption{Performance of the proposed framework variants in terms of modality combination on Kinetics400.}    \vspace*{-2mm}
		\begin{tabular}{cccc}
			\bottomrule
			Scheme & Modality & Top-1 (\%) & Top-5 (\%)\\
			\hline
			\multirow{2}{1.5cm}{\centering Uni-modal}& Audio&$ 26.8 $ &$ 45.1 $ \\
			& RGB&$ 79.1 $ &$ 93.9 $  \\
			\hline
			\multirow{3}{1.5cm}{\centering Bi-modal}& Audio-Flow&$ 56.7 $ &$ 72.8 $ \\
			& Audio-RGB&$ 81.6 $ &$ 94.8 $  \\
			& RGB-Flow&$ 81.1 $ &$ 94.3 $  \\
			\hline
			\multirow{1}{1.6cm}{\centering Multi-modal}& Audio-RGB-Flow&$ 82.8 $&$ 95.7 $ \\
			\toprule
		\end{tabular}
		\label{tab:Modality}
	\end{table}
	%\subsubsection{Trainable IMD}
	\subsubsection{Trainable Irrelevant Modality Dropout (IMD)}
	\noindent\textbf{Overlapping Loss.} 
	The overlap between the output predictions of the pre-trained audio model AST and the corresponding $ K $ relevant textual audio labels in the SAVLD is computed as ground truth for the RN optimization. The overlap is then normalized between $ 0 $ and $ 1 $ to match the sigmoidal prediction of RN. We tested two overlap loss functions IOU and Dice overlap \cite{diceloss}. The upper part of Table \ref{tab:IMD} shows the performance of two functions. The overlap function does not show a large effect on the framework performance because this loss is performed only to build the ground truth for the binary cross-entropy rather than to optimize the framework itself.
	
	\noindent\textbf{Most $k$ Audio Relevant Labels.}
	The overlap between AST predictions and the relevant textual label vectors is computed with same size of two vectors. However, we tested the impact of using different sizes of $k$ against $Y_a$. Table \ref{tab:IMD} shows that $10-20$ is the best $k-Y_a$ ratio for ground truth computation.

	\noindent\textbf{Relevance Thresholding ($\alpha$).}
	The sigmoidal predictions of the RN indicate the audio relevance to the visual modality on the input video sample. Thus, these sigmoidal predictions range between 0 and 1, which are going to be used by the thresholding layer in the dropout module. The effect of various threshold values on the dropout rate affecting the framework performance is reported in Table \ref{tab:IMD}. Notably, when $\alpha=0.25$, the framework shows better results.
	\begin{table}
		\fontsize{9}{12}\selectfont
		\centering   
		\caption{Impact of different IMD settings on the proposed framework performance on Kinetics400.} \vspace*{-2mm}
		\begin{tabular}{lccccc}
			\bottomrule
			%		Loss & K &$\alpha$&M/W& Top-1 & Top-5 \\
			Loss & $K$-$Y_a$ &$\alpha$&M/W& Top-1 (\%)& Top-5 (\%)\\
			\hline
			\textbf{IOU} &$ 10$-$10 $&$ 0.50 $&M&$ 81.7 $&$ 95.4 $\\
			\textbf{Dice} &$ 10$-$10 $&$ 0.50 $&M&$ 81.6 $&$ 95.5 $\\
			\hline
			IOU & \textbf{20-10} &$ 0.50 $&M&$ 81.7 $&$ 94.8 $\\
			IOU & \textbf{10-20} &$ 0.50 $&M&$ 82.3 $&$ 95.5 $\\
			IOU & \textbf{20-20} &$ 0.50 $&M&$ 81.9 $&$ 95.1 $\\
			\hline
			IOU &$ 10$-$20 $& \textbf{0.25} &M& \textbf{82.8} &\textbf{95.7}\\
			IOU &$ 10$-$20 $& \textbf{0.75} &M&$ 80.9 $&$ 94.4 $\\
			\hline
			IOU &$ 10$-$20 $&$ 0.25 $&\textbf{W}&$ 80.4 $&$ 94.3 $\\
			\toprule
		\end{tabular}
		
		\label{tab:IMD}
	\end{table}
	
	\noindent\textbf{Irrelevant Modality Masking Scheme.} Two types of masking schemes in IMD are evaluated. The first scheme is the default explained masking (\textbf{M}) method in this work in which the relevant sigmoidal score larger than $\alpha$ is simply multiplied with the audio embeddings received from the audio Transformer, whereas the relevance is less than $\alpha$ is multiplied by $0$ to completely mask the audio embeddings from the fusion step. The second scheme is applying the same weighting (\textbf{W}) method in two cases when relevant sigmoidal scores are larger or less than $\alpha$. The lower part of Table \ref{tab:IMD} shows that the weighting relevance method provides less performance compared with the masking method.
	\subsection{Comparison with State-of-the-art}
	
	%\subsubsection{Modality Fusion Methods}
	\noindent\textbf{Modality Fusion Methods.} We compare our proposed IMD to other fusion methods in the literature. When we adopt the relevant fusion methods, we disable the RN learning to ensure the fusion process is performed on each audio-visual input sample. Several fusion methods, including SE Gate \cite{UnimodalBetterMultimodal}, NL Gate \cite{UnimodalBetterMultimodal}, late-concat \cite{UnimodalBetterMultimodal}, AVC \cite{AVC}, and GMU \cite{GatedUnit}, are evaluated. Table \ref{tab:fusion} reports the performance of our framework when using each method of these fusion methods.
	\begin{table}[!h]
		\fontsize{9}{11}\selectfont
		\centering   
		\caption{Performance comparison of the proposed IMD against other relevant fusion methods on Kinetics400. IMD$^*$ refers to apply the proposed intra-class cross-modality augmentation.} \vspace*{-2mm}
		\begin{tabular}{lcc}
			\bottomrule %\hline
			Fusion Method & Top-1  (\%)& Top-5  (\%)\\
			\hline
			SE Gate \cite{UnimodalBetterMultimodal}&$ 79.2 $&$ 94.1 $\\
			NL Gate \cite{UnimodalBetterMultimodal}&$ 80.8 $&$ 94.8 $\\
			late-concat \cite{UnimodalBetterMultimodal} &$ 78.6 $&$ 93.9 $\\
			AVC \cite{AVC}& $ 79.5 $&$ 94.2 $\\
			GMU \cite{GatedUnit}&$ 80.9 $&$ 94.8 $\\
			\hline
			\textbf{IMD (ours)} & \textbf{82.3} &\textbf{95.1}\\
			\textbf{IMD$^*$(ours)} & \textbf{82.8} &\textbf{95.7}\\
			\toprule
		\end{tabular}
		
		\label{tab:fusion}
	\end{table}
	
	\begin{table}
		\fontsize{9}{11}\selectfont
		\centering   
		\caption{Comparison to prior uni and multimodal action recognition methods on Kinetics400 dataset. Most compared methods are Transformer-based except the CNN-based methods: MoViNet-A5, SlowFast, and X3D. All methods are organized by their size in which the upper part involves the Transformers of base (\textbf{B}) size, whereas the lower part involves the ones with large (\textbf{L}) size.} \vspace*{-2mm}
		\begin{tabular}{lcccr}%{cccc} %{@{}lc@{}lc@{}}
			\bottomrule
			Model & P-train &Top-1 & Top-5 & GFLOPs\\
			\toprule
			MoViNet-A5 \cite{MoViNets}&IN-1K&$ 78.2 $& N/A&$ 29 $ \\
			SlowFast 16x8 \cite{slowfast}&IN-1K&$ 79.8 $&$ 93.9 $&$ 7,020 $  \\
			VATT-B \cite{DropToken}&IN-1K&$ 79.6 $&$ 94.9 $ &$ 9,090 $ \\
			X3D-XXL \cite{x3d}&IN-1K&$ 80.4 $&$ 94.6 $ &$ 5,823 $ \\
			TimeSFormer-B \cite{timesformerT}&IN-21K& $ 78.0 $ &$ 93.7 $&$ 590 $\\
			Swin-B \cite{swinT}&IN-1K&$ 80.6 $& $ 94.6 $& $ 3,384 $\\
			Swin-B \cite{swinT}&IN-21K&$ 82.7 $& $ 95.5 $& $ 3,384 $\\
			X-ViT-16x \cite{XViT}&IN-21K& $ 80.2 $&$ 94.7 $&$ 850 $ \\
			\textbf{IMD-B (ours)} & IN-21K &\textbf{82.8}&\textbf{95.7} & $ 4,464 $\\
			\hline%\bottomrule
			%\rowcolor{LightBlue}\hline
			AVSlowFast \cite{AVSlowFast}&-&$ 78.8 $&$ 93.6 $&$ 7,020 $  \\
			G-Blend \cite{UnimodalBetterMultimodal}&Sport1M&$ 80.4 $&$ 94.8 $ & $ 3,303 $\\%\rowcolor{LightBlue}
			MBT (AV) \cite{BottleneckFusion}&IN-21K&$ 80.8 $&$ 94.6 $ & N/A\\
			\hline%\bottomrule
			TimeSFormer-L \cite{timesformerT}&IN-21K& $ 80.7 $ &$ 94.7 $&$ 7,140 $\\
			Mformer-L \cite{Mformer} &IN-1K& $ 80.2 $& $ 94.8 $&$ 35,553 $\\
			ViViT-L \cite{viviT}&IN-21K&$ 80.6 $&$ 94.7 $& $ 17,352 $ \\
			VATT-L \cite{DropToken}&IN-1K&$ 82.1 $&$ 95.5 $ &$ 29,800 $ \\
			Swin-L \cite{swinT}&IN-21K&$ 83.1 $& $ 95.9 $& $ 7,248 $\\
			\textbf{IMD-L (ours)} & IN-21K &\textbf{84.2}&\textbf{96.3} & $ 8,232 $\\
			\toprule
		\end{tabular}
		
		\label{tab:Kinetics400}
	\end{table}
	\begin{table}
		\fontsize{9}{11}\selectfont
		\centering   
		\caption{Comparison to to prior multimodal action recognition methods on UCF-101. Methods are organized by pretrained data.} \vspace*{-2mm}
		\begin{tabular}{lcc}
			\bottomrule
			Model & Pretrained Weights & Top-1 (\%) \\
			\toprule
			%		AVSlowFast \cite{AVSlowFast}&-& $ 77.4 $ \\
			CoViAR \cite{compressedvideo}&IN-1K& $ 90.4 $ \\
			TSN \cite{tsn}&IN-1K& $ 94.2 $ \\
			CoViAR + OF \cite{compressedvideo}&IN-1K& $ 94.9 $ \\
			Two-stream fusion \cite{TwoStream2016Fusion}&IN-1K& $ 92.5 $ \\
			MM-ViT \cite{MultimodalCompressedVideo}&IN-21K&$  95.4 $ \\
			\textbf{IMD-B (ours)} &IN-21K&\textbf{95.9}\\        
			\hline
			I3D \cite{kinetics400}&Kinetics& $ 95.4 $ \\
			MM-ViT \cite{MultimodalCompressedVideo}&Kinetics& $ 98.9 $ \\
			\textbf{IMD-B (ours)} &Kinetics&\textbf{99.1}\\
			\toprule
		\end{tabular}
		
		\label{tab:UCF-101}
	\end{table}
	%\subsubsection{Action Recognition}
	\noindent\textbf{Action Recognition.} In this part, we evaluate our overall framework against the state-of-the-art multimodal methods for video-based action recognition on two datasets Kinetics400 \cite{kinetics400} and UCF-101 \cite{ucf101}. Several uni/multimodal action recognition methods have shown good performance including the CNN-based models: TSN \cite{tsn}, I3D \cite{kinetics400}, AVSlowFast \cite{AVSlowFast}, SlowFast \cite{slowfast}, X3D \cite{x3d}, MoViNet-A5 \cite{MoViNets}, and G-Blend \cite{UnimodalBetterMultimodal} along with the Transformer-based models: VATT \cite{DropToken}, TimeSFormer \cite{timesformerT}, Swin Video Transformer \cite{swinT}, X-ViT \cite{XViT}, MBT \cite{BottleneckFusion}, Mformer \cite{Mformer}, ViViT-L \cite{viviT}, MM-ViT \cite{MultimodalCompressedVideo}. However, our multimodal Transformer-based framework outperforms all aforementioned methods on the two datasets with considerable margins. Table \ref{tab:Kinetics400} reports the action recognition performance on Kinetics400 dataset, and Table \ref{tab:UCF-101} reports the results on UCF-101 dataset. Unlike our framework, most multimodal video-based deep models are usually evaluated on multimodality datasets. However, we compared our method with the best available models on these modality-specific datasets in which our IMD models provides a Top-1 performance boost of the most similar multimodal methods (highlighted in LightBlue color) G-Blend and MBT with $\sim$$3.8\%$ and $\sim$$3.4\%$, respectively.
	
	\subsection{Discussion and Limitations}
	On the basis of the reported results and overall framework building, few worth-mentioning points need to be declared. \textit{First}, our framework is a convolution-free Transformer-based neural network. It shows a larger performance enhancement on large datasets compared with its performance on small datasets, as shown in Tables \ref{tab:Kinetics400} and \ref{tab:UCF-101}. \textit{Second}, our designed visual-modality two-stream Transformer attempts to cut the number of Transformer parameters while preserving the best spatiotemporal representation learning. \textit{Third}, audio-modality masking based on the sigmoidal scores provided by the RN is relayed on the audio Transformer accuracy on its audio dataset (\eg, AudioSet). For example, as reported in Table \ref{tab:IMD}, the best $\alpha=0.25$. This finding is because the AST and most audio recognition models in the literature still do not show as high recognition accuracy as visual-based models because audio modality is more challenging compared with visual data. AST \cite{AST} provides mAP of $0.485$. In this manner, we think that threshold $\alpha$ can be higher when a better pretrained audio model is used. \textit{Fourth}, the proposed intra-class cross-modality augmentation method shows a positive effect on the framework training. However, its main effectiveness is related to the framework design, which considers that the audio modality is noisy and unlabeled. Thus, intra-class cross-modality augmentation improves the performance by pairing visual and auditory modalities. \textit{Finally}, even though we evaluate the quality of the SAVDL dictionary indirectly using its impact on the action recognition performance. However, it is of future interest to build a new evaluation method that can effectively evaluate the quality of text-to-text similarity matching.  
	
	\section{Conclusion}
	In this work, we introduced a novel audio-visual action recognition framework. It leverages the power of NLP and audio recognition models to automatically annotate audio modality in the input unlabeled audio for video-based action recognition in an online training manner. The proposed framework features a novel IMD that completely drops the irrelevant audio modality based on a recommendation provided by an RN, which in turn learns to estimate the audio modality relevance to the input video category. Our framework aims to create cross-modality high semantic features that can help to bridge the relevance gap between the visual and auditory modalities in videos. As long as a strong pre-trained model on a large dataset is available to be used as a transfer learning task, this training scheme can be adopted by other bimodality and multimodality models.  Thus, the same idea can be further investigated for other modalities, such as textual-auditory and textual-visual model training, which is part of our future interest.\\
	
	%\section*{Acknowledgement}
	\noindent\textbf{Acknowledgments}
	\noindent\small{This work was supported in part by the National Natural Science Foundation of China under grants U21A20455, 61972265, 61872429 and 11871348, the Natural Science Foundation of Guangdong Province of China under grant 2020B1515310008, Educational Commission of Guangdong Province of China under grant 2019KZDZX1007, and the Shenzhen Basis Research Project under grant JCYJ20180305125521534.}

	%%%%%%%%% REFERENCES
	{\small
		\bibliographystyle{unsrt}
		%\bibliography{egbib}
		\bibliography{MultimodalCVPRalfasly}

\begin{thebibliography}{10}

\bibitem{BottleneckFusion}
Arsha Nagrani, Shan Yang, Anurag Arnab, Aren Jansen, Cordelia Schmid, and Chen
  Sun.
\newblock Attention bottlenecks for multimodal fusion.
\newblock In A.~Beygelzimer, Y.~Dauphin, P.~Liang, and J.~Wortman Vaughan,
  editors, {\em Advances in Neural Information Processing Systems}, 2021.

\bibitem{fusion}
Olga Zatsarynna, Yazan Abu~Farha, and Juergen Gall.
\newblock Multi-modal temporal convolutional network for anticipating actions
  in egocentric videos.
\newblock In {\em CVPR Workshops}, pages 2249--2258, June 2021.

\bibitem{EfficientMultiModalTransformer}
Sangho Lee, Youngjae Yu, Gunhee Kim, Thomas Breuel, Jan Kautz, and Yale Song.
\newblock Parameter efficient multimodal transformers for video representation
  learning.
\newblock In {\em International Conference on Learning Representations}, 2021.

\bibitem{twostreamMulitmodal}
Lu~Chi, Guiyu Tian, Yadong Mu, and Qi~Tian.
\newblock {Two-Stream Video Classification with Cross-Modality Attention}.
\newblock In {\em ICCVW}, pages 4511--4520. IEEE, oct 2019.

\bibitem{MultimodalCompressedVideo}
Jiawei Chen and Chiu~Man Ho.
\newblock {MM-ViT: Multi-Modal Video Transformer for Compressed Video Action
  Recognition}.
\newblock {\em arXiv}, aug 2021.

\bibitem{GatedUnit}
John Arevalo, Thamar Solorio, Manuel Montes-Y-G{\'{o}}mez, and Fabio~A.
  Gonz{\'{a}}lez.
\newblock {Gated multimodal units for information fusion}.
\newblock {\em 5th International Conference on Learning Representations, ICLR
  2017 - Workshop Track Proceedings}, 2019.

\bibitem{UnimodalBetterMultimodal}
Weiyao Wang, Du~Tran, and Matt Feiszli.
\newblock {What Makes Training Multi-Modal Classification Networks Hard?}
\newblock {\em Proceedings of the IEEE Computer Society Conference on Computer
  Vision and Pattern Recognition}, pages 12692--12702, 2020.

\bibitem{AVC}
Relja Arandjelovic and Andrew Zisserman.
\newblock {Look, Listen and Learn}.
\newblock In {\em ICCV}, volume Octob, pages 609--617, oct 2017.

\bibitem{EPIC-KITCHENS}
Dima Damen, Hazel Doughty, Giovanni~Maria Farinella, Sanja Fidler, Antonino
  Furnari, Evangelos Kazakos, Davide Moltisanti, Jonathan Munro, Toby Perrett,
  Will Price, and Michael Wray.
\newblock Scaling egocentric vision: The epic-kitchens dataset.
\newblock In {\em European Conference on Computer Vision (ECCV)}, 2018.

\bibitem{kinetics400}
Joao Carreira and Andrew Zisserman.
\newblock {Quo Vadis, Action Recognition? A New Model and the Kinetics
  Dataset}.
\newblock In {\em CVPR}, pages 4724--4733. IEEE, jul 2017.

\bibitem{ucf101}
Khurram Soomro, Amir~Roshan Zamir, and Mubarak Shah.
\newblock {UCF101: A Dataset of 101 Human Actions Classes From Videos in The
  Wild}.
\newblock {\em arXiv}, dec 2012.

\bibitem{AudioSet}
Jort~F. Gemmeke, Daniel~P.W. Ellis, Dylan Freedman, Aren Jansen, Wade Lawrence,
  R.~Channing Moore, Manoj Plakal, and Marvin Ritter.
\newblock {Audio Set: An ontology and human-labeled dataset for audio events}.
\newblock {\em ICASSP, IEEE International Conference on Acoustics, Speech and
  Signal Processing - Proceedings}, pages 776--780, 2017.

\bibitem{VGGsound}
Honglie Chen, Weidi Xie, Andrea Vedaldi, and Andrew Zisserman.
\newblock {VGGSOUND : A large-scale audio-visual dataset}.
\newblock In {\em ICASSP}, pages 721--725, 2020.

\bibitem{BERT}
Jacob Devlin, Ming-Wei Chang, Kenton Lee, Kristina~Toutanova Google, and A~I
  Language.
\newblock {BERT: Pre-training of Deep Bidirectional Transformers for Language
  Understanding}.
\newblock {\em Naacl-Hlt 2019}, 2018.

\bibitem{GloVe}
Paul~M. Brennan, James~J.M. Loan, Neil Watson, Pragnesh~M. Bhatt, and
  Peter~Alwyn Bodkin.
\newblock {Pre-operative obesity does not predict poorer symptom control and
  quality of life after lumbar disc surgery}.
\newblock {\em British Journal of Neurosurgery}, 31(6):682--687, 2017.

\bibitem{DropToken}
Hassan Akbari, Liangzhe Yuan, Rui Qian, Wei-Hong Chuang, Shih-Fu Chang, Yin
  Cui, and Boqing Gong.
\newblock {VATT: Transformers for Multimodal Self-Supervised Learning from Raw
  Video, Audio and Text}.
\newblock In {\em NeurIPS}, 2021.

\bibitem{2streams2014}
Karen Simonyan and Andrew Zisserman.
\newblock {Two-Stream Convolutional Networks for Action Recognition in Videos}.
\newblock {\em Advances in Neural Information Processing Systems}, jun 2014.

\bibitem{TwoStream2016Fusion}
Christoph Feichtenhofer, Axel Pinz, and Andrew Zisserman.
\newblock {Convolutional Two-Stream Network Fusion for Video Action
  Recognition}.
\newblock {\em Proceedings of the IEEE Computer Society Conference on Computer
  Vision and Pattern Recognition}, 2016-Decem(i):1933--1941, 2016.

\bibitem{MMTM}
Hamid Reza~Vaezi Joze, Amirreza Shaban, Michael~L. Iuzzolino, and Kazuhito
  Koishida.
\newblock {MMTM: Multimodal transfer module for CNN fusion}.
\newblock {\em CVPR}, pages 13286--13296, 2020.

\bibitem{VLM}
Hu~Xu, Gargi Ghosh, Po-yao Huang, Prahal Arora, Masoumeh Aminzadeh, Christoph
  Feichtenhofer, Florian Metze, and Luke Zettlemoyer.
\newblock {VLM: Task-agnostic Video-Language Model Pre-training for Video
  Understanding}.
\newblock In {\em Findings of the Association for Computational Linguistics:
  ACL-IJCNLP 2021}, pages 4227--4239, Stroudsburg, PA, USA, 2021. Association
  for Computational Linguistics.

\bibitem{ObjectsthatSound}
Relja Arandjelovi{\'{c}} and Andrew Zisserman.
\newblock {Objects that Sound}.
\newblock In {\em European Conference on Computer Vision (ECCV)}, 2018.

\bibitem{hardModalityDropout}
Ahmed {Hussen Abdelaziz}, Barry~John Theobald, Paul Dixon, Reinhard Knothe,
  Nicholas Apostoloff, and Sachin Kajareker.
\newblock {Modality Dropout for Improved Performance-driven Talking Faces}.
\newblock {\em ICMI 2020 - Proceedings of the 2020 International Conference on
  Multimodal Interaction}, pages 378--386, 2020.

\bibitem{fusion2}
Alejandro Cartas, Jordi Luque, Petia Radeva, Carlos Segura, and Mariella
  DImiccoli.
\newblock {Seeing and hearing egocentric actions: How much can we learn?}
\newblock {\em ICCVW}, pages 4470--4480, 2019.

\bibitem{EPICFusion}
Evangelos Kazakos, Arsha Nagrani, Andrew Zisserman, and DIma Damen.
\newblock {EPIC-Fusion: Audio-Visual Temporal Binding for Egocentric Action
  Recognition}.
\newblock In {\em ICCV}, volume 2019-Octob, pages 5491--5500. IEEE, oct 2019.

\bibitem{compressedvideo}
Chao-Yuan Wu, Manzil Zaheer, Hexiang Hu, R.~Manmatha, Alexander~J. Smola, and
  Philipp Krahenbuhl.
\newblock {Compressed Video Action Recognition}.
\newblock In {\em CVPR}, pages 6026--6035. IEEE, jun 2018.

\bibitem{CluesLabels}
Yu~Wu and Yi~Yang.
\newblock Exploring heterogeneous clues for weakly-supervised audio-visual
  video parsing.
\newblock In {\em CVPR}, 2021.

\bibitem{ViT}
Alexey Dosovitskiy, Lucas Beyer, Alexander Kolesnikov, Dirk Weissenborn,
  Xiaohua Zhai, Thomas Unterthiner, Mostafa Dehghani, Matthias Minderer, Georg
  Heigold, Sylvain Gelly, Jakob Uszkoreit, and Neil Houlsby.
\newblock An image is worth 16x16 words: Transformers for image recognition at
  scale.
\newblock In {\em International Conference on Learning Representations}, 2021.

\bibitem{Alayrac2020}
Jean~Baptiste Alayrac, Adri{\`{a}} Recasens, Rosalia Schneider, Relja
  Arandjelovi{\'{c}}, Jason Ramapuram, Jeffrey de~Fauw, Lucas Smaira, Sander
  Dieleman, and Andrew Zisserman.
\newblock {Self-supervised multimodal versatile networks}.
\newblock {\em Advances in Neural Information Processing Systems},
  2020-Decem(ii):1--19, 2020.

\bibitem{Owens2018}
Andrew Owens and Alexei~A. Efros.
\newblock {Audio-Visual Scene Analysis with Self-Supervised Multisensory
  Features}.
\newblock In {\em ECCV}, volume 11210 LNCS, pages 639--658, 2018.

\bibitem{dropout}
Srivastava Nitish, Hinton Geoffrey, Krizhevsky Alex, Sutskever Ilya, and
  Salakhutdinov Ruslan.
\newblock {Dropout: A Simple Way to Prevent Neural Networks from Overfitting}.
\newblock {\em Journal of Machine Learning Research}, 2014.

\bibitem{ModDrop}
Natalia Neverova, Christian Wolf, Graham Taylor, and Florian Nebout.
\newblock {ModDrop: Adaptive Multi-Modal Gesture Recognition}.
\newblock {\em IEEE Transactions on Pattern Analysis and Machine Intelligence},
  38(8):1692--1706, aug 2016.

\bibitem{gatemodality}
Feiran Huang, Kaimin Wei, Jian Weng, and Zhoujun Li.
\newblock {Attention-Based Modality-Gated Networks for Image-Text Sentiment
  Analysis}.
\newblock {\em ACM Transactions on Multimedia Computing, Communications, and
  Applications}, 16(3):1--19, sep 2020.

\bibitem{mixup}
Hongyi Zhang, Moustapha Cisse, Yann~N. Dauphin, and David Lopez-Paz.
\newblock {MixUp: Beyond empirical risk minimization}.
\newblock In {\em International Conference on Learning Representations(ICLR)},
  2018.

\bibitem{MLSL-imagenet}
Olga Russakovsky, Jia Deng, Hao Su, Jonathan Krause, Sanjeev Satheesh, Sean Ma,
  Zhiheng Huang, Andrej Karpathy, Aditya Khosla, Michael Bernstein,
  Alexander~C. Berg, and Li~Fei-Fei.
\newblock {ImageNet Large Scale Visual Recognition Challenge}.
\newblock {\em IJCV}, 115:211--252, 2015.

\bibitem{timesformerT}
Gedas Bertasius, Heng Wang, and Lorenzo Torresani.
\newblock Is space-time attention all you need for video understanding?
\newblock In {\em ICML}, volume 139, pages 813--824, Jul 2021.

\bibitem{selfAttention}
Ashish Vaswani, Noam Shazeer, Niki Parmar, Jakob Uszkoreit, Llion Jones,
  Aidan~N. Gomez, {\L}ukasz Kaiser, and Illia Polosukhin.
\newblock {Attention is all you need}.
\newblock In {\em Proceedings of the 31st International Conference on Neural
  Information Processing Systems}, pages 6000--6010, California, USA, 2017. ,.

\bibitem{LN}
Jimmy~Lei Ba, Jamie~Ryan Kiros, and Geoffrey~E. Hinton.
\newblock {Layer Normalization}.
\newblock {\em arXiv}, 2016.

\bibitem{AST}
Yuan Gong, Yu-An Chung, and James Glass.
\newblock {AST: Audio Spectrogram Transformer}.
\newblock {\em arXiv}, 2021.

\bibitem{pytorchh}
Adam Paszke, Sam Gross, Francisco Massa, Adam Lerer, James Bradbury, Gregory
  Chanan, Trevor Killeen, Zeming Lin, Natalia Gimelshein, Luca Antiga, Alban
  Desmaison, Andreas K{\"{o}}pf, Edward Yang, Zach DeVito, Martin Raison,
  Alykhan Tejani, Sasank Chilamkurthy, Benoit Steiner, Lu~Fang, Junjie Bai, and
  Soumith Chintala.
\newblock {PyTorch: An imperative style, high-performance deep learning
  library}.
\newblock In {\em Advances in Neural Information Processing Systems}, 2019.

\bibitem{word2vec}
Tomas Mikolov, Ilya Sutskever, Kai Chen, Greg Corrado, and Jeffrey Dean.
\newblock {Distributed Representations of Words and Phrases and their
  Compositionality}.
\newblock In {\em NIPS}, oct 2013.

\bibitem{diceloss}
W.R. Crum, Oscar Camara, and D.L.G. Hill.
\newblock {Generalized Overlap Measures for Evaluation and Validation in
  Medical Image Analysis}.
\newblock {\em IEEE Transactions on Medical Imaging}, 25(11):1451--1461, nov
  2006.

\bibitem{MoViNets}
Dan Kondratyuk, Liangzhe Yuan, Yandong Li, Li~Zhang, Mingxing Tan, Matthew
  Brown, and Boqing Gong.
\newblock {MoViNets: Mobile Video Networks for Efficient Video Recognition}.
\newblock pages 16020--16030, 2021.

\bibitem{slowfast}
Christoph Feichtenhofer, Haoqi Fan, Jitendra Malik, and Kaiming He.
\newblock {Slowfast networks for video recognition}.
\newblock In {\em ICCV}, 2019.

\bibitem{x3d}
Christoph Feichtenhofer.
\newblock {X3D: Expanding Architectures for Efficient Video Recognition}.
\newblock {\em CVPR}, pages 200--210, 2020.

\bibitem{swinT}
Ze~Liu, Jia Ning, Yue Cao, Yixuan Wei, Zheng Zhang, Stephen Lin, and Han Hu.
\newblock {Video Swin Transformer}.
\newblock {\em arXiv}, pages 1--12, 2021.

\bibitem{XViT}
Adrian Bulat, Juan-Manuel Perez-Rua, Swathikiran Sudhakaran, Brais Martinez,
  and Georgios Tzimiropoulos.
\newblock Space-time mixing attention for video transformer.
\newblock In {\em NeurIPS}, 2021.

\bibitem{AVSlowFast}
Fanyi Xiao, Yong~Jae Lee, Kristen Grauman, Jitendra Malik, and Christoph
  Feichtenhofer.
\newblock {Audiovisual SlowFast Networks for Video Recognition}.
\newblock {\em arXiv}, 2020.

\bibitem{Mformer}
Mandela Patrick, Dylan Campbell, Yuki~M. Asano, Ishan Misra, Florian Metze,
  Christoph Feichtenhofer, Andrea Vedaldi, and Jo{\~{a}}o~F. Henriques.
\newblock {Keeping Your Eye on the Ball: Trajectory Attention in Video
  Transformers}.
\newblock In {\em Advances in Neural Information Processing Systems}, jun 2021.

\bibitem{viviT}
Anurag Arnab, Mostafa Dehghani, Georg Heigold, Chen Sun, Mario
  Lu{\v{c}}i{\'{c}}, and Cordelia Schmid.
\newblock {ViViT: A Video Vision Transformer}.
\newblock {\em arXiv}, 2021.

\bibitem{tsn}
Limin Wang, Yuanjun Xiong, Zhe Wang, Yu~Qiao, Dahua Lin, Xiaoou Tang, and Luc
  {Van Gool}.
\newblock {Temporal Segment Networks: Towards Good Practices for Deep Action
  Recognition}.
\newblock In {\em ECCV 2016}, pages 20--36. 2016.

\bibitem{hmdb51}
Hilde Kuehne, Hueihan Jhuang, E.~Garrote, T.~Poggio, and T.~Serre.
\newblock {HMDB: A large video database for human motion recognition}.
\newblock In {\em ICCV}, pages 2556--2563. IEEE, nov 2011.

\end{thebibliography}
	}
\pagebreak
%\noindent------------------------------------------------------------------------------\\\\
\noindent\textbf{\Large Supplementary File}\\\\
%\noindent\textbf{{\large Further Results and Analysis}}\\\\
%\section{Further Results and Analysis}
\noindent\textbf{\small Semantic Audio-Video Label Dictionary (SAVLD)} \vspace*{2mm}

\begin{figure*}%[H]
	\centering
	\includegraphics[trim=0 25 0 0, width=17.5cm,clip, keepaspectratio]{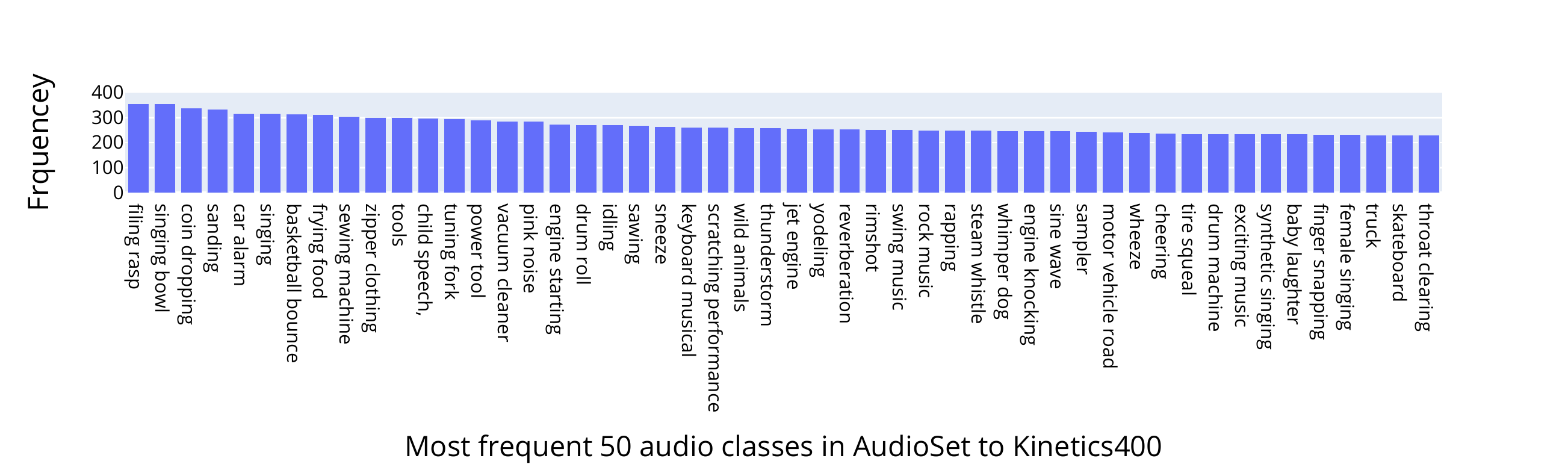}
	%	\vspace*{-3mm}
	\caption{The most 50 frequent audio labels in AudioSet mapped to Kinetics400 labels when $k=50$.}
	\label{fig:MostKinetics400}
\end{figure*}

\begin{figure*}%[!h]
	\centering
	\includegraphics[trim=0 0 0 0, width=17.5cm,clip, keepaspectratio]{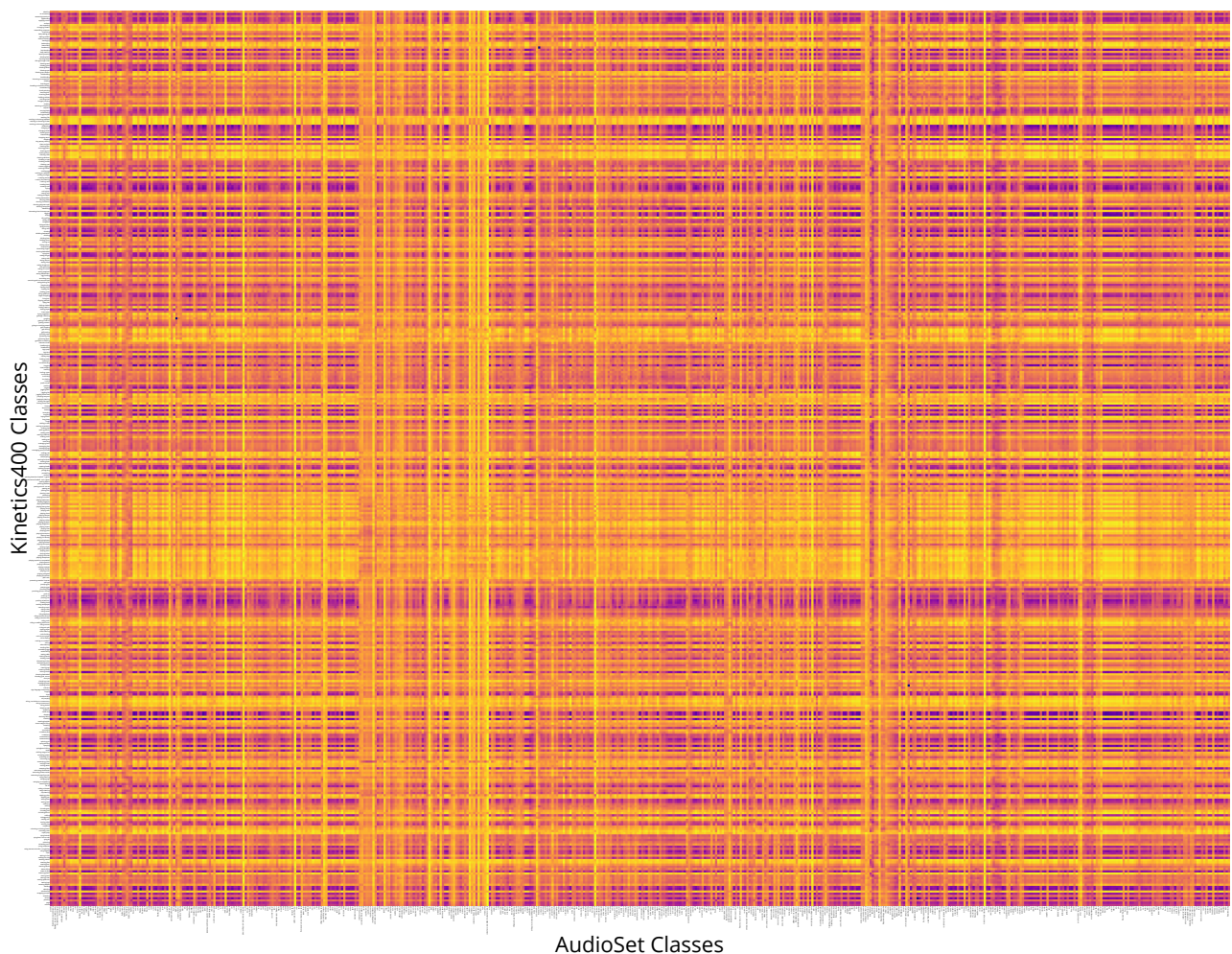}
	%	\vspace*{-3mm}
	\caption{The heatmap of the semantic relevance estimated by BERT between Kinetics400 labels and AudioSet labels. The darker the more relevant.}
	\label{fig:Kinetics400}
\end{figure*}

\begin{figure*}%[!h]
	\centering
	\includegraphics[trim=0 25 0 0, width=17.5cm,clip, keepaspectratio]{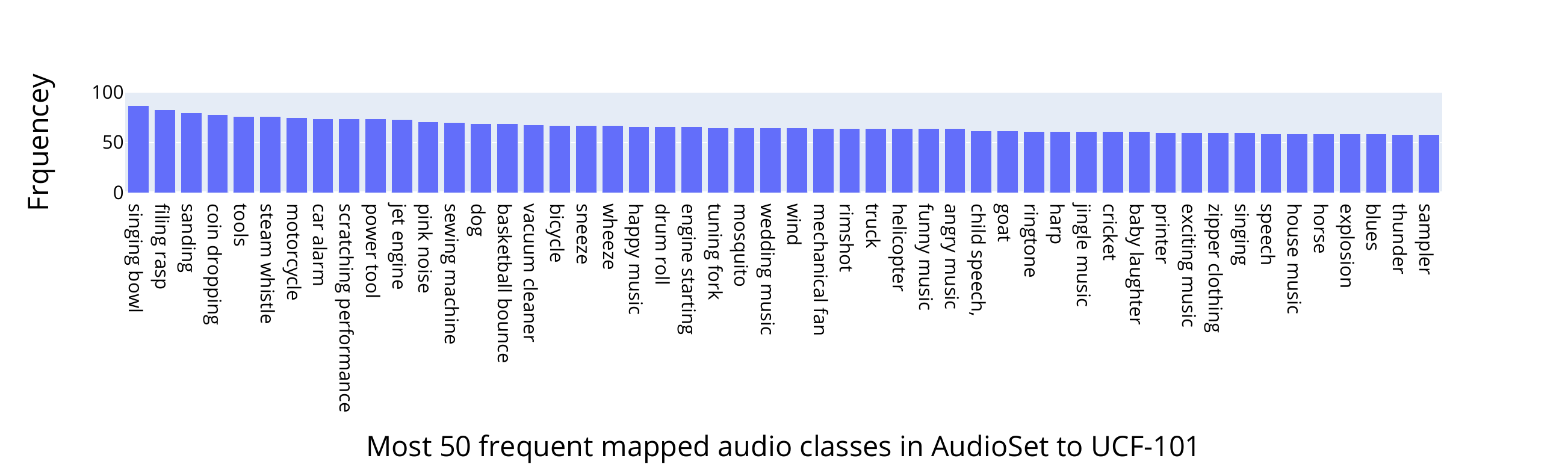}
	%	\vspace*{-3mm}
	\caption{The most 50 frequent audio labels in AudioSet mapped to UCF-101 labels when $k=50$.}
	\label{fig:MostUCF}
\end{figure*}

\begin{figure*}%[!h]
	\centering
	\includegraphics[trim=0 0 0 0, width=17.5cm,clip, keepaspectratio]{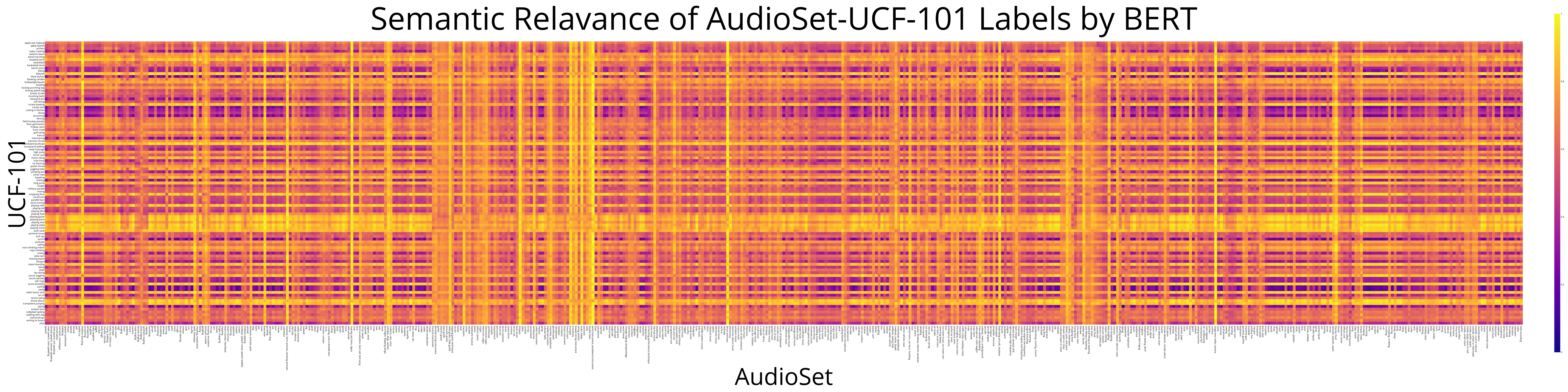}
	%	\vspace*{-3mm}
	\caption{The heatmap of the semantic relevance estimated by BERT between UCF-101 labels and AudioSet labels. The darker the more relevant.}
	\label{fig:UCF}
\end{figure*}

For building the SAVLD, we use \textbf{\textit{BERT-base uncased}} in which the label text is lowercased before tokenizing it. Therefore, as a preprocessing step, we first normalized all textual labels into lowercase form. Notably, the resulting label dictionary SAVLD is not very accurate since the semantic gap between video and audio datasets is still large. It is also because the number of classes on video/audio datasets is still considered small (\ie Kinetics400 has $400$ labels and AudioSet has $ 527 $ label). However, using SAVLD dictionaries, our framework narrows the input noise by using the audio predictions. Then, it matches the audio scene to its closest similar visual class regardless of the fact that many audio classes do not have large relevance. In the training phase, the labeling process is guided by applying the IOU function between the AST predictions and the corresponding audio labels to the video label in the SAVLD dictionary. Table \ref{tab:SAVD} shows a small part of the Kinetics400-AudioSet label dictionary when $k=5$. Additionally, we visualize the AudioSet labels' relevance to Kinetics400 and UCF-101 labels in Fig. \ref{fig:Kinetics400} and Fig. \ref{fig:UCF}, respectively. The most frequent mapped audio labels to the vision-specific dataset labels are shown in Fig. \ref{fig:MostKinetics400} and Fig. \ref{fig:MostUCF} for Kinetics400 and UCF-101 datasets, respectively.\footnote{We implemented our framework in Pytorch in which we borrowed several parts from the following codebases:\\
	https://github.com/huggingface/transformers
	https://github.com/YuanGongND/ast\\
	https://github.com/facebookresearch/SlowFast\\
	https://github.com/facebookresearch/TimeSformer\\
	https://github.com/rwightman/pytorch-image-models
	https://github.com/unixpickle/audioset\\
	https://github.com/ekazakos/temporal-binding-network\\
	https://github.com/marl/l3embedding\\
	https://github.com/johnarevalo/gmu-mmimdb
}\\\\
%\subsection{Experiments on HMDB51 and Kinetics-Sounds}
\noindent\textbf{\small Experiments on HMDB51 and Kinetics-Sounds} \vspace*{2mm}

We further evaluate our method against relevant methods for video-based action recognition in two more datasets HMDB51 \cite{hmdb51} and Kinetics-Sounds \cite{ucf101}. \textbf{HMDB51} \cite{hmdb51} is split into three overlapped splits. It contains $6,766$ videos of $51$ classes with an average length of $3$ seconds. \textbf{Kinetics-Sounds} \cite{AVC} is a subset from the original Kinetics400 \cite{kinetics400} dataset. It contains $34$ classes in which each class videos have a remarkable sound signature. Since Kinetics dataset editions are downloadable from YouTube, its size may vary by time as some videos may get removed. Herein, we use $19,627$ videos for training and $1,344$ videos for evaluation. length of $3$ seconds. The average video length in this dataset is $10$ seconds.

\begin{table}
	\fontsize{9}{11}\selectfont
	\centering
	\caption{Performance comparison to relevant visual-based methods (RGB + Optical Flow) on HMDB51 dataset.}
	\begin{tabular}{l|c}
		\bottomrule
		Model &  Top-1 (\%) \\
		\toprule
		CoViAR  \cite{compressedvideo}& $ 59.1 $ \\
		Two-stream fusion \cite{TwoStream2016Fusion}& $ 65.4 $ \\
		TSN \cite{tsn}& $ 69.4 $ \\
		I3D \cite{kinetics400}& $ 66.4 $ \\		
		CoViAR + OF \cite{compressedvideo}& $ 70.2 $ \\
		\textbf{IMD-B (ours)} &\textbf{71.3}\\
		\toprule
	\end{tabular}
	
	\label{tab:hmdb}
\end{table}

\begin{table}
	\fontsize{9}{11}\selectfont
	\centering	
	\caption{Performance comparison to relevant methods on Kinetics-Sounds dataset.}
	\begin{tabular}{l|c|c}
		\bottomrule
		Model & P-train & Top-1 (\%) \\
		\toprule
		L3-Net \cite{AVC}&IN-1K& $ 74.0 $ \\
		SlowFast R101 \cite{slowfast}&IN-1K& $ 77.9 $ \\
		AVSlowFast, R101 \cite{AVSlowFast}&IN-1K& $ 85.0 $ \\
		MBT (AV) \cite{BottleneckFusion}&IN-21K& $ 85.0 $ \\
		\textbf{IMD-B (ours)} &IN-21K&\textbf{90.48}\\
		\textbf{IMD-B$^\star$ (ours)} &IN-21K&\textbf{91.44}\\
		\toprule
	\end{tabular}
	
	\label{tab:UCF-101}
\end{table}
\begin{table*}
	\centering	
	\caption{Samples of the most relevant AudioSet labels to video labels retrieved by semantic sentence-based embeddings mapping by BERT when $K=5$.}
	\begin{tabular}{cll} 
		\bottomrule
		Dataset & Label & Relevant AudioSet Labels\\
		\toprule
		\parbox[t]{1mm}{\multirow{19}{*}{\rotatebox[origin=c]{90}{Kinetics400}}} & playing guitar & bass guitar;acoustic guitar;guitar;chopping food;electric guitar\\
		&applauding & speech;applause;whistling;chime;clapping\\
		&belly dancing & rapping;yodeling;synthetic singing;child singing;frying food\\
		&canoeing or kayaking & rowboat, canoe, kayak;motorboat, speedboat;skateboard;folk music;sailboat, sailing ship\\
		&clean and jerk & fill with liquid;pump liquid;filing rasp;rumble;rustle\\
		&country line dancing & female singing;dance music;male singing;salsa music;drum roll\\
		&driving car & emergency vehicle;motor vehicle road;filing rasp;car;engine starting\\
		&feeding birds & wild animals;insect;mosquito;bird;patter\\
		&gargling & gargling;gurgling;snoring;reversing beeps;yodeling\\
		&kissing & whispering;typing;cheering;laughter;breathing\\
		&pumping gas & frying food;pump liquid;sawing;sanding;filing rasp\\
		&recording music & vocal music;music;soundtrack music;wedding music;jingle music\\
		&scrambling eggs & singing bowl;spray;thunder;wheeze;tools\\
		&sniffing & whimper;growling;cheering;whispering;rattle \\
		&sneezing & gurgling;snoring;babbling;gargling;rapping \\
		&tickling & whispering;rustle;cheering;growling;screaming\\
		&yawning & babbling;rapping;frying food;gurgling;snoring \\
		&writing & writing;speech;typing;chatter;mechanisms \\
		&whistling & whistling;humming;whistle;whip;siren\\
		&welding & gears;scissors;drill;boiling;bicycle\\
		\toprule
	\end{tabular}
	
	\label{tab:SAVD}
\end{table*}
Table \ref{tab:hmdb} reports the performance of several visual action recognition methods including CoViAR  \cite{compressedvideo}, Two-stream fusion \cite{TwoStream2016Fusion}, TSN \cite{tsn}, I3D \cite{kinetics400}, and CoViAR + OF \cite{compressedvideo}. Notably, our method provides a slight performance boost on HMDB51 because our method is a Transformer-based framework that shows better improvement on large datasets in terms of number of videos per class. However, our framework provides top-1 of $71.3\%$ which is better compared with CoViAR + OF \cite{compressedvideo} with a performance boost of $\sim$$1.1\%$.
We compare our method with several methods on the visual-audio annotated dataset Kinetics-Sounds. This dataset was first used in \cite{AVC} as a subset of the main Kinetics400 dataset \cite{kinetics400}. Our framework provides a significant boost in this dataset, where it provides top-1 of $90.48\%$ and $91.44\%$ with and without intra-class cross-modality augmentation, respectively. Since this dataset is an audio-video annotated in which audio and video are mostly relevant in each video, the cross-modality augmentation does not improve the performance. This interprets our finding regarding this augmentation method, where it provides most performance boosts on datasets with low audio-video relevance. Therefore, in our method, cross-modality augmentation is particularly applied for improving the video-based human activity recognition on vision-specific videos. \\\\

%\subsection{Visual Two-Stream Transformer Variants}
\noindent\textbf{\small The Visual Two-Stream Transformer} \vspace*{2mm}

In this part, we report the performance of three Transformer variants of the proposed two-stream visual Transformer. In order to leverage the pretrained ImageNet knowledge, we have adopted several parts from ViT \cite{ViT} in terms of number of Transformer encoder blocks, embedding size, number of self-attention heads, input dimensions. Our Transformer scales properties are almost similar to the \cite{ViT}, \ie \textit{small}, \textit{base}, and \textit{large}, except the spatiotemporal encoder blocks. We used one spatiotemporal block on the \textit{small} encoder instance as it involves $8$ blocks, whereas we add $2$ and $4$ spatiotemporal blocks for \textit{base} and \textit{large} instances as they involve $12$ and $24$ blocks, respectively. Table \ref{table:encoder} reports the recognition performance on Kinetics400 dataset as well as the Transformer instances' costs in terms of number of parameters and GFLOPs. 
\begin{table*}
	\fontsize{9}{11}\selectfont
	\begin{center}
		\renewcommand{\arraystretch}{1.2}
		\caption{Performance of the proposed visual two-stream Transformer with different size. Three Transformer instances are trained, where each of which is initialized with its ViT corresponding ImageNet-21K weights. The performance is reported on Kinetics400. Number of parameters is reported in million.}
		\begin{tabular}{c|c|c|c|c}
			\bottomrule
			T. Size & Top-1 (\%)  &	Top-5 (\%)& Params & GFLOPs \\
			\hline
			Small &$78.8$&	$92.8$&$47.9$$\times$$2$& $2,871$ \\
			Base &$81.1$&	$94.3$&	$88.6$$\times$$2$&$4,464$ \\
			Large&$82.6$&	$95.2$&$173.7$$\times$$2$ &$8,232$\\\toprule
		\end{tabular}		
		
		\label{table:encoder}
	\end{center}
\end{table*}

\end{document}